\documentclass{article}

\usepackage[nonatbib,preprint]{neurips_2019}
 \usepackage{neurips_2019}

\usepackage[utf8]{inputenc} 
\usepackage[T1]{fontenc}    
\usepackage{hyperref}       
\usepackage{url}            
\usepackage{booktabs}       
\usepackage{amsfonts}       
\usepackage{nicefrac}       
\usepackage{microtype}      

\usepackage{amsmath}
\usepackage{graphicx}

\title{Local Unsupervised Learning for Image Analysis}

\author{
  Leopold Grinberg \\
  IBM Research\\
  \texttt{lgrinbe@us.ibm.com} \\
   \And
  John Hopfield \\
  Princeton Neuroscience Institute  \\
  Princeton University \\
  \texttt{hopfield@princeton.edu} \\
  \And  
  Dmitry Krotov \\
  MIT-IBM Watson AI Lab \\
  IBM Research\\
  \texttt{krotov@ibm.com} \\
}

\begin{document}

\maketitle

\begin{abstract}
  Local Hebbian learning is believed to be inferior in performance to end-to-end training using a backpropagation algorithm. We question this popular belief by designing a local algorithm that can learn convolutional filters at scale on large image datasets. These filters combined with patch normalization and very steep non-linearities result in a good classification accuracy for shallow networks trained locally, as opposed to end-to-end. The filters learned by our algorithm contain both orientation selective units and unoriented color units, resembling the responses of pyramidal neurons located in the cytochrome oxidase ``interblob’’ and ``blob’’ regions in the primary visual cortex of primates. It is  shown that convolutional networks with patch normalization significantly outperform standard convolutional networks on the task of recovering the original classes when shadows are superimposed on top of standard CIFAR-10 images. Patch normalization approximates the retinal adaptation to the mean light intensity, important for human vision. We also demonstrate a successful transfer of learned representations between CIFAR-10 and ImageNet $32\times 32$ datasets. All these results taken together hint at the possibility that local unsupervised training might be a powerful tool for learning general representations (without specifying the task) directly from unlabeled data. 
\end{abstract}

\section{Introduction}
Local learning, motivated by Hebbian plasticity, is usually believed to be inferior in performance to gradient-based optimization, for example a backpropagation algorithm. Common arguments include the following ideas. Feature detectors in the early layers of neural networks should be specifically crafted to solve the task that the neural network is designed for, thus some information about the errors of the network or the loss functions must be available during learning of the early layer weights. Making random changes in weights and accepting those changes that improve accuracy, as evolutionary algorithms do, is very inefficient in large networks. Gradient-based optimization seems to converge to the desired solution faster than alternative methods, which do not rely on the gradient.

At the same time, modern neural networks are heavily overparametrized, which means that there are many combinations of weights that lead to a good generalization performance. Thus, there is an appealing idea that this manifold of ``good'' weights might be found by a purely local learning algorithm that operates directly on the input data without the information about the output of the network or the task to be performed.

A variety of local learning algorithms relying only on bottom-up propagation of the information in neural networks have been recently discussed in the literature \cite{Chklovskii, Pehlevan, Hawkins, Seung, Bahroun, Krotov_Hopfield_2019}. A recent paper \cite{Krotov_Hopfield_2019}, for example, proposed a learning algorithm that is local and unsupervised in the first layer. It manages to learn useful early features necessary to achieve a good generalization performance, in line with networks trained end-to-end on simple machine learning benchmarks. The limitation of this study is that the proposed algorithms were tested only in fully connected networks, only on pixel permutation invariant tasks, and only on very simple datasets: pixel permutation invariant MNIST and CIFAR-10. An additional limitation of study \cite{Krotov_Hopfield_2019}, which is not addressed in this work, is that they studied neural networks with only one hidden layer.

Our main contributions are the following. Based on the open source implementation \cite{bio_learning_github} of the algorithm proposed in \cite{Krotov_Hopfield_2019},  we designed an unsupervised learning algorithm for networks with {\em local} connectivity. We wrote a fast CUDA library that allows us to quickly learn weights of the convolutional filters at scale. We propose a modification to the standard convolutional layers, which includes patch normalization and very steep non-linearities in the activation functions. These unusual architectural features together with the proposed learning algorithm allow us to match the performance of networks of similar size and architecture trained using the backpropagation algorithm end-to-end on CIFAR-10. On  ImageNet $32\times 32$ the accuracy of our algorithm is slightly worse than the accuracy of the network trained end-to-end, but it's in the same ballpark.  The usefulness of patch normalization is illustrated by designing an artificial test set from CIFAR-10 images that are dimmed by shadows. The network with patch normalization outperforms the standard convolutional network by a large margin on this task. At the end, transfer learning between CIFAR-10 and ImageNet $32 \times 32$ is discussed.

The main goal of our work is to investigate the concept of local bottom-up learning and its usefulness for machine learning and generalization, and not to design a biologically plausible framework for deep learning. We acknowledge that the proposed algorithm uses shared weights in convolutional filters, which is not biological. Despite this lack of the overall biological motivation in this work, we note that the filters learned by our algorithm show a well pronounced separation between color-sensitive cells and orientation-selective cells. This computational aspect of the algorithm matches nicely with a similar separation between the stimulus specificity of the responses of neurons in  blob and interblob pathways, known to exist in the V1 area of the visual cortex. 

\section{Learning Algorithm and Network Architecture}
During training, each input image is cut into small patches of size $W\times W\times 3$. Stride of $1$ pixel and no padding is used at this stage. The resulting set of patches $v_i^A$ (index $i$ enumerates pixels, index $A$ enumerates different patches) is shuffled at each epoch and is organized into minibatches that are presented to the learning algorithm. The learning algorithm uses weights $M_{\mu i}$, which is a matrix of $K$ channels by $N=W\cdot W\cdot 3$ visible units, that are initialized from a standard normal distribution and iteratively updated according to the following learning rule \cite{Krotov_Hopfield_2019}
\begin{equation}
\Delta M_{\mu i} = \varepsilon \sum\limits_{A\in \text{minibatch} }g\Big[ \text{Rank}\Big(\sum\limits_j M_{\mu j} v_j^A\Big) \Big] \Big[ v_i^A- \Big(\sum\limits_k M_{\mu k} v_k^A\Big) M_{\mu i} \Big]\label{learning rule}
\end{equation}
where $\varepsilon$ is the learning rate. The activation function $g(\cdot)$ is equal to one for the strongest driven channel and is equal to a small negative constant for the channel that has a rank $m$ in the activations 
\begin{equation}
g(i) = \left\{ 
\begin{array}{cl}1, & \text{if}\ i=1\\ 
-\Delta, & \text{if }\ i=m\\
0, & \text{otherwise}
\end{array}\right.\label{discrete learning activation function}
\end{equation}
Ranking is done for each element of the minibatch separately. The weights are updated after each minibatch for a certain number of epochs, until each row of the weight matrix converges to a unit vector. This is a literal implementation of the algorithm \cite{Krotov_Hopfield_2019} adapted to small patches of images.

The resulting matrix $M_{\mu i}$ is used as weights of the convolutional filters with two important modifications. Frist, each patch $v_i$ of the image is normalized to be a unit vector before taking the dot product with the weight matrix. Given that the rows of the weight matrix themselves are unit vectors, the dot product between the weight and the patch is a cosine of the similarity between the two. Thus, $\sum_iM_{\mu i } v_i \in [-1, 1 ]$. Second, the result of the dot product is passed through a very steep non-linearity - rectified power function \cite{DAM2016, DAM2018}
\begin{equation}
f(x) = \Big[ \text{ReLU}(x) \Big]^n 
\end{equation}
where the power $n$ is a hyperparameter of that layer. We call these slightly unusual convolutional layers NNL-CONV layers (normalized non-linear convolutional layers), in order to distinguish them from the standard ones, denoted CONV in this paper. The standard CONV layers do not use per-patch normalization and use ReLU as an activation function. In the following sections we also use standard max-pooling layers and standard fully connected layers with softmax activation function for the classifier.

\section{Evaluation of the model on CIFAR-10 dataset}\label{section CIFAR}
\begin{figure}[h]
\begin{center}
\includegraphics[width = 0.6\linewidth]{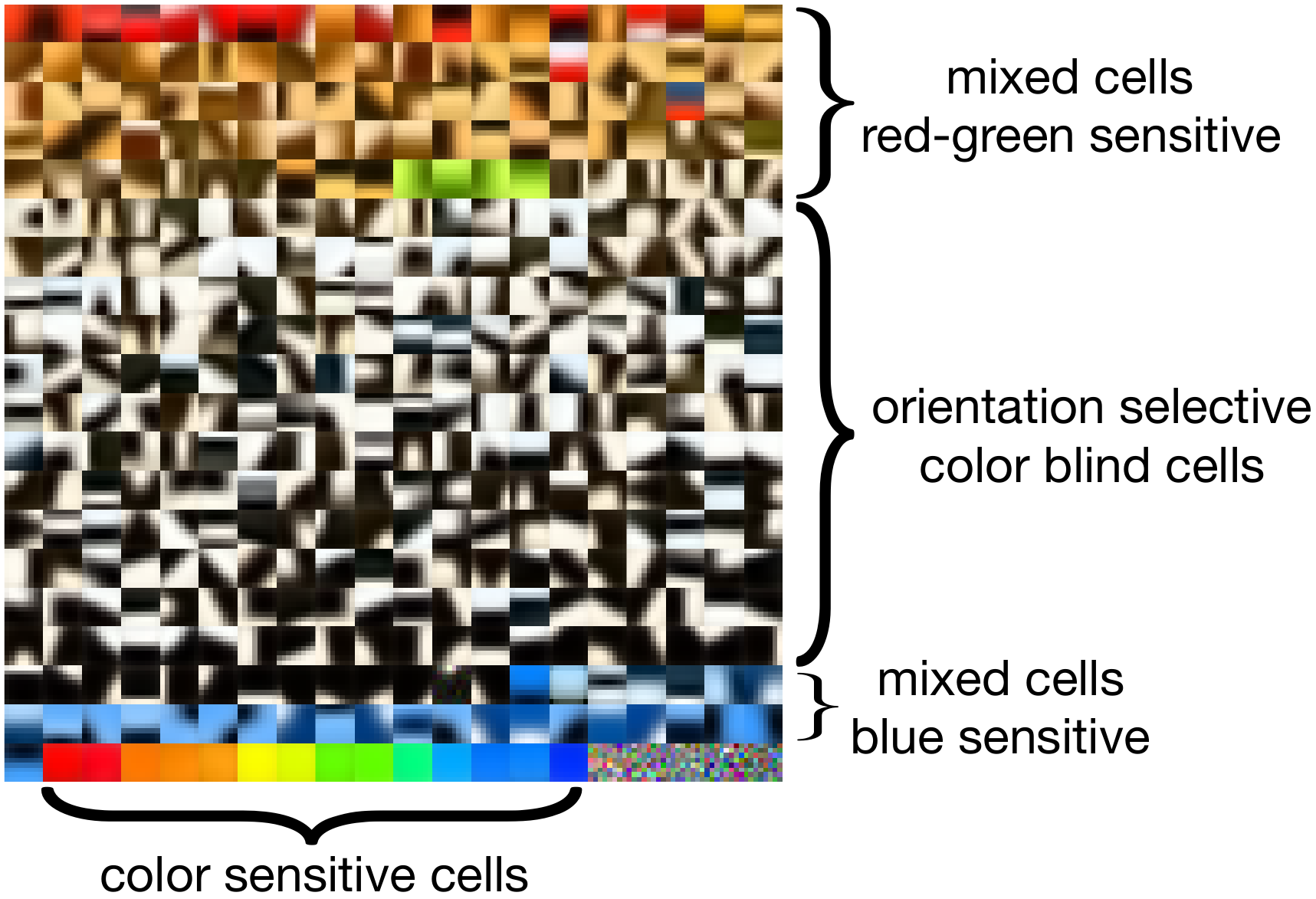}
\end{center}
\caption{\footnotesize{Filters of size $W=8$ pixels, trained using the proposed algorithm (\ref{learning rule}). The filters are ordered to emphasize the differences between several groups of cells. The last five units in the last row did not learn any useful representation. They can be deleted from the network.}}\label{bio receptive fields}
\end{figure}

The algorithm (\ref{learning rule}) was applied to images from CIFAR-10 dataset to learn the weights of the NNL-CONV filters. They are shown in Fig.\ref{bio receptive fields}. Each small square corresponds to a different channel (hidden unit) and is shown by projecting its corresponding weights into the image plane $W\times W\times 3$. The connection strength to each pixel has $3$ components corresponding to the RGB color image. These weights are linearly stretched so that they change on a segment $[0,1]$ for each channel. Thus, black color corresponds to synaptic weights that are either equal to zero, or negative; white color corresponds to weights that are large and have approximately equal values in all three RGB channels; blue color corresponds to weights that have large weights connected to blue neurons, and small or zero weights connected to red and green neurons, etc. 
As is clear from this figure, the resulting weights show a diversity of features of the images, including line detectors, color detectors, and detectors of more complicated shapes.  These feature detectors appear to be much smoother than the feature detectors resulting from standard end-to-end training of CNNs with the backpropagation algorithm. 

Guided by these examples, one can make the following qualitative observations:
\begin{enumerate}
\item The majority of the hidden units are black and white, having no preference for R,G, or B color. These units respond strongly to oriented edges, corners, bar ends, spots of light, more complicated shapes, etc.
\item There is a significant presence of hidden units detecting color. Those neurons tend to have a smaller preference for orientation selectivity, compared to black and white units. 
\end{enumerate}
  
In order to test the quality of these learned filters with respect to the generalization performance, they were used as the weights of a simple network with one NNL-CONV layer. 
\begin{figure}[h]
\begin{center}
\includegraphics[width = 1.0\linewidth]{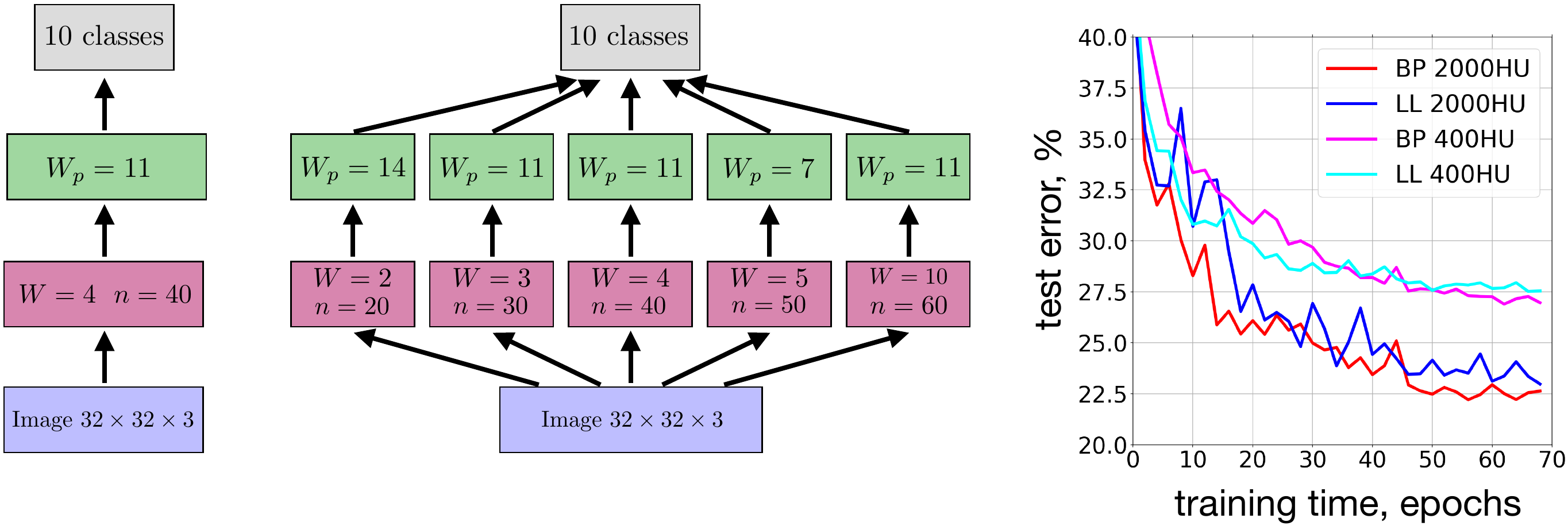}
\end{center}
\caption{\footnotesize{(Left) A simple architecture with an RGB image projected to the NNL-CONV layer (shown in red); window size $W=4$; the power of the activation function is $n=40$; number of channels $K=400$ (all the hyperparameters were determined on the validation set, see Appendix for details). This convolutional layer is followed by a max-pooling layer, shown in green;  pooling window size $W_p=11$ pixels. The last layer is a softmax classifier. (Middle) An architecture with blocks of different $W$. Power $n$, determined on the validation set, is shown in each NNL-CONV block. (Right) The test set error of the two networks shown in the left and middle panels together with their end-to-end counterparts as a function of supervised training time. }}\label{networks errors}
\end{figure} 
Consider for example the architecture shown in Fig. \ref{networks errors} (left). In this architecture a NNL-CONV layer is followed by a max pooling layer and then a fully connected softmax classifier. The weights of the NNL-CONV layer were fixed, and given by the output of the algorithm (\ref{learning rule}). The max-pooling layer does not have any trainable weights. The weights of the top layer were trained using the gradient decent based optimization (Adam optimizer with the cross-entropy loss function). The accuracy of this network was compared with the accuracy of the network of the same capacity, with NNL-CONV layers replaced by the standard CONV layers, trained end-to-end using Adam optimizer. The results are shown in Fig. \ref{networks errors} (right). Here one can see how the errors on the held-out test set decrease as training progresses. Training time here refers to the training of the top layer classifier only in the case of the NNL-CONV network, and training of all the layers (convolutional layer and the  classifier) in the case of the standard CONV network trained end-to-end. For the simple network shown on the left the error of the locally trained network is $27.80\%$, the error of the network trained end-to-end is $27.11\%$.

In order to achieve a better test accuracy it also helps to organize the NNL-CONV layer as a sequence of blocks with various sizes of the window $W$, like in Fig. \ref{networks errors} (middle). Having this diversity of the receptive windows allows the neural network to detect features of different scales in the input images. Same as above, the performance of this network trained using algorithm (\ref{learning rule}) is compared with the performance of a similar size network trained end-to-end. The results are shown by the red and blue curves in Fig.  \ref{networks errors} (right). The error of the locally trained network is $23.40\%$, the error of the network trained end-to-end is $22.57\%$.

The main conclusion here is that the networks with filters obtained using the local unsupervised algorithm (\ref{learning rule}) achieve almost the same accuracy as the networks trained end-to-end. This result is at odds with the common belief that the first layer feature detectors should be crafted specifically to solve the narrow task specified by the top layer classifier. Instead, this suggests that a general set of weights of the first layer can be learned using the local bottom-up unsupervised training (\ref{learning rule}), and that it is sufficient to communicate the task only to the top layer classifier, without adjusting the first layer weights.

We have done an extensive set of experiments with networks of different size of the hidden layer as well as with different sizes of convolutional windows $W$, pooling windows $W_p$, different strides in NNL-CONV and pooling layers, and different powers $n$ of the activation function. The conclusions are the following: 
\begin{enumerate}
\item The classification accuracy increases as the hidden layer gets wider. This is a known phenomenon for networks trained end-to-end with the backpropagation algorithm. The same phenomenon is valid for the networks trained using the local learning (\ref{learning rule}). 
\item For a given choice of windows $W$ of the blocks of the NNL-CONV layer the remaining hyperparameters (powers $n$, pooling windows $W_p$, strides, size of the minibatch, etc) can be optimized on the validation set (see Appendix for details) so that the accuracy of the locally trained network is almost the same as the accuracy of the network with the same set of windows $W$ trained end-to-end.
\end{enumerate}

In all the experiments in this section a standard validation procedure was used. Standard CIFAR-10 training set of 50000 images was randomly split into 45000 training images and 5000 validation images. The hyperparameters were adjusted to optimize the accuracy on the validation set. Once this is done, the validation set was combined together with the training set to retrain the models for the  optimal values of the hyperparameters. These models were tested on the standard held-out test set of 10000 images.

We also acknowledge that better accuracies on CIFAR-10 dataset are achievable \cite{CIFAR accuracies}, but they require the use of one (or several) architectural/algorithmic methods that go beyond the limits of this work, such as: deeper architectures, dropout, data augmentation, injection of noise during training, etc.

\section{Evaluation of the model on ImageNet $32 \times 32$ dataset}
\begin{figure}[h]
\begin{center}
\includegraphics[width = 0.8\linewidth]{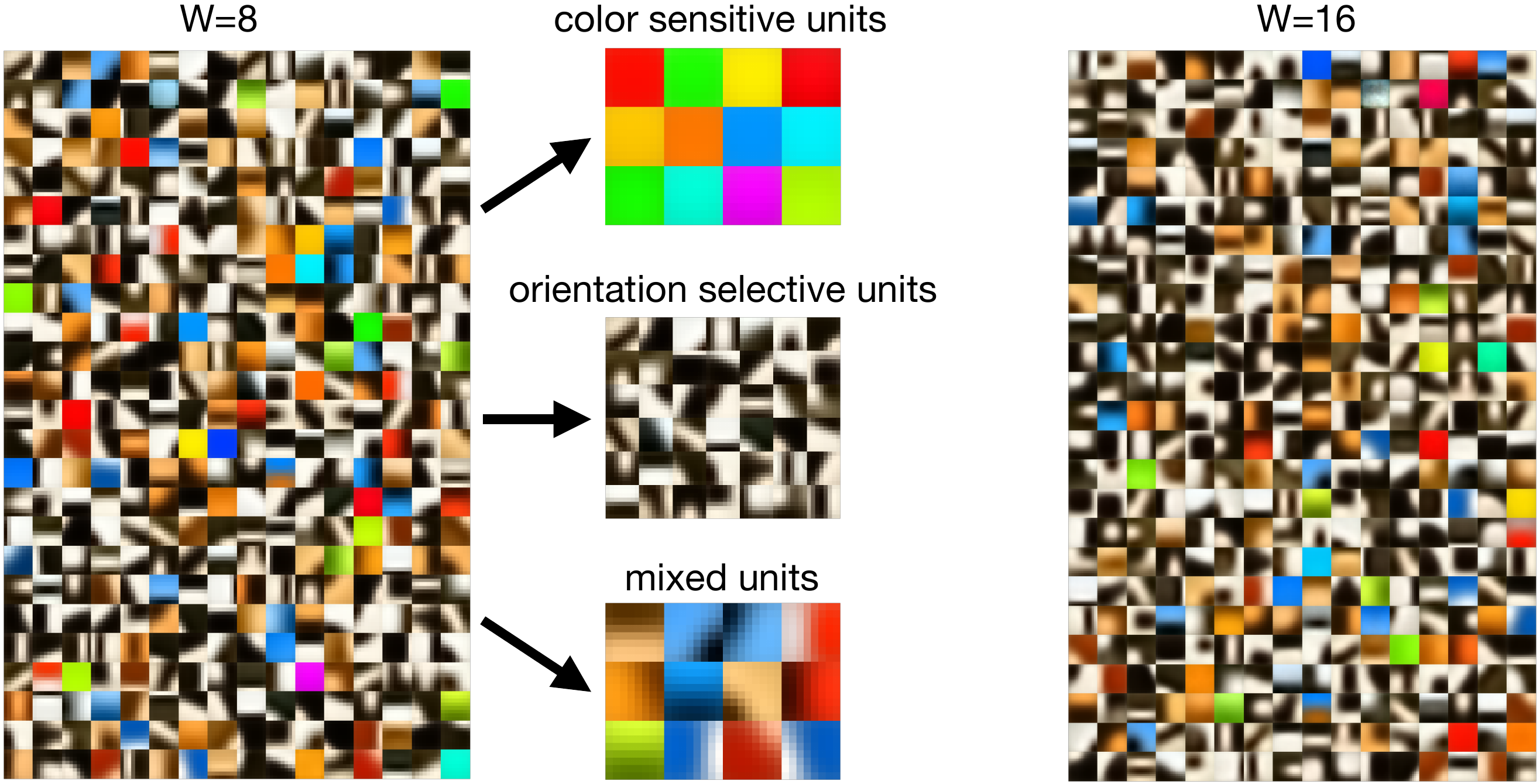}
\end{center}
\caption{\footnotesize{Convolutional filters of two sizes $W=8$ and $W=16$ trained on ImageNet $32\times32$. Examples of color sensitive units, orientation selective units and mixed units are shown in the middle of the figure. In the color sensitive examples the first and the fourth units in the top row look very similar - both are red. Although this might be difficult to see, the shades of those two red colors are different.}}\label{ImageNet_rec_fields_fig}
\end{figure}

A similar set of experiments were conducted on ImageNet $32 \times 32$ dataset \cite{ImageNet32}. A large family of filters with windows $W$ changing in the range $2\leq W \leq 16$ were trained using the algorithm (\ref{learning rule}).  Examples of those learned filters are shown in Fig. \ref{ImageNet_rec_fields_fig}.  Visually, they look similar to the filters resulting from training on CIFAR-10, see Fig. \ref{bio receptive fields}. The separation between color sensitive and orientation selective cells is present for all sizes of windows $W$, as is the case for CIFAR-10.  

In order to benchmark the accuracy of the models $10\%$ of the standard training set was used as a validation set to tune the hyperparameters. After the hyperparameters were chosen, the validation set was combined with the training set to retrain the models for the optimal values of the hyperparameters. The accuracy was measured on the standard held-out test set.

\begin{figure}[h]
\begin{center}
\includegraphics[width = 0.8\linewidth]{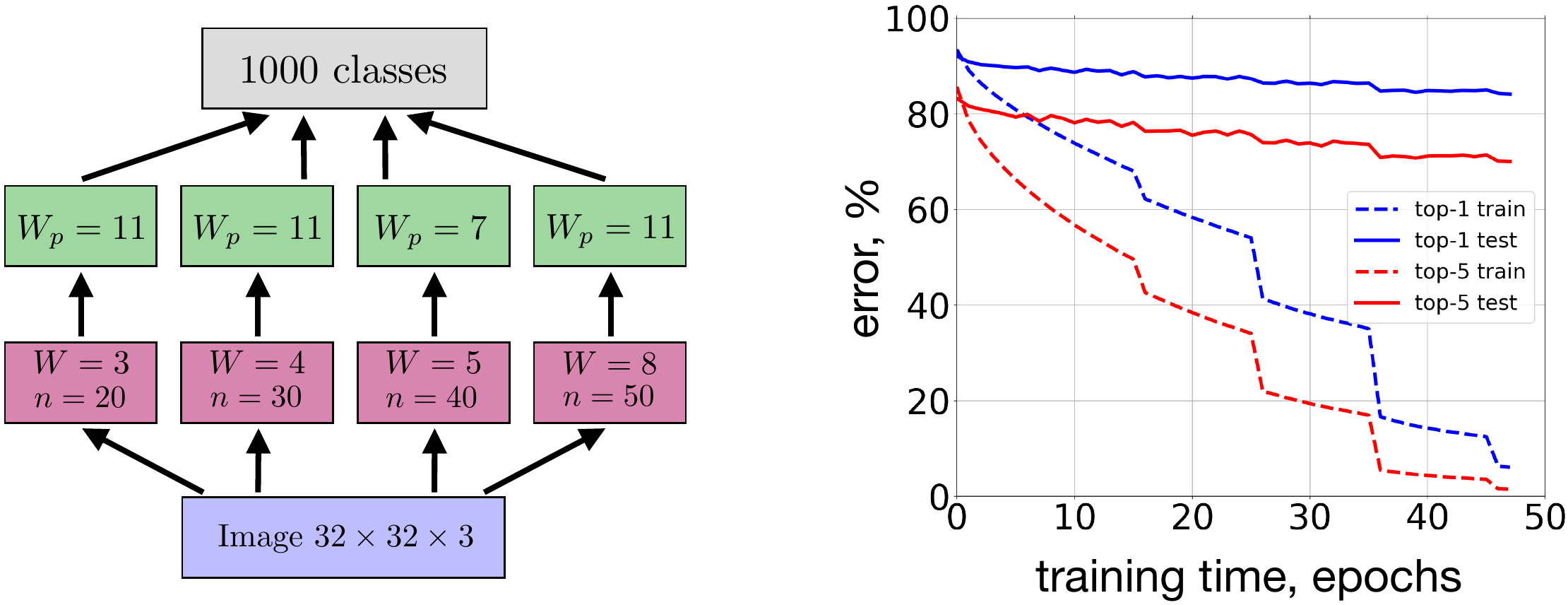}
\end{center}
\caption{\footnotesize{The architecture of the network used for ImageNet $32\times 32$ experiments together with the errors on the training and test set for the locally trained network. See Appendix for the complete list of the hyperparameters. }}\label{network_ImageNet}
\end{figure}

A family of models with one layer of NNL-CONV units, having block structure was considered. The ``optimal'' model is shown in Fig. \ref{network_ImageNet}
and consists of four blocks with windows of sizes $W=3,4,5,8$ pixels. The error rate of the model is: $84.13\%$ in top-1 classification, and $70.00\%$ in top-5 classification.  The errors on the training and test sets are shown in Fig. \ref{network_ImageNet}. Details of these experiments are in the Appendix.  For comparison, the network of the same size trained end-to-end achieves: top-1 error $79.72\%$, top-5 error $62.58\%$. Thus, on this task the locally trained network performs slightly worse than the network trained end-to-end, but the difference is not that big. Especially considering that no information about the class labels was used in training the NNL-CONV filters. We also acknowledge that significantly better accuracies can be achieved on this dataset by training CONV networks end-to-end, see table 1 in  \cite{ImageNet32}, but this requires deep architectures.

\section{Color Sensitivity, Orientation Selectivity and Cytochrome Oxidase Stain}
The filters shown in Fig. \ref{bio receptive fields}, \ref{ImageNet_rec_fields_fig} are unit vectors, as a result of convergence of learning rule (\ref{learning rule}), and are multiplied by pixel intensities of a patch, which is also normalized to be  a unit vector. Thus, the maximal value of this dot product is equal to one, when the patch matches exactly the filter (assuming that the filter does not have negative elements). Thus, the learned filters can be interpreted as preferred stimulus (input images which maximize firing) of the corresponding hidden units. Below we describe a metaphorical comparison of these learned preferred stimuli with preferred stimuli of cells in the V1 area of the visual cortex in primates.

An interesting anatomical feature of the primary visual cortex is revealed when cells are stained with a cytochrome oxidase enzyme. This staining shows a pattern of blobs and interblob regions. In the famous set of experiments \cite{Livingstone Hubel} it was discovered that the cells in the interblob regions are highly orientation selective and respond to luminance rather than color, while the neurons inside the blobs respond to colors and are insensitive to orientation.

There are many important details and subtleties of these experiments that are not discussed in this paper, e.g. the  neocortical visual areas of the primate brain have six anatomical layers many clearly divided into sublayers;  neurons in different layers have different response properties;  the response properties also vary smoothly within a layer.  These  facts limit the usefulness of describing how a ``typical'' pyramidal cell in V1 responds, especially to strong natural stimuli. However, the literature supports the qualitative assertion that there exist a segregation of orientation selective and color processing cells \cite{Livingstone Hubel, Johnson}. A segregation resembling the one discussed above can be seen in the preferred stimuli of the hidden neurons learned by our algorithm, see Fig. \ref{bio receptive fields}, \ref{ImageNet_rec_fields_fig}.

\begin{figure}[h]
\vspace{-0.2cm}
\begin{center}
\includegraphics[width = 0.6\linewidth]{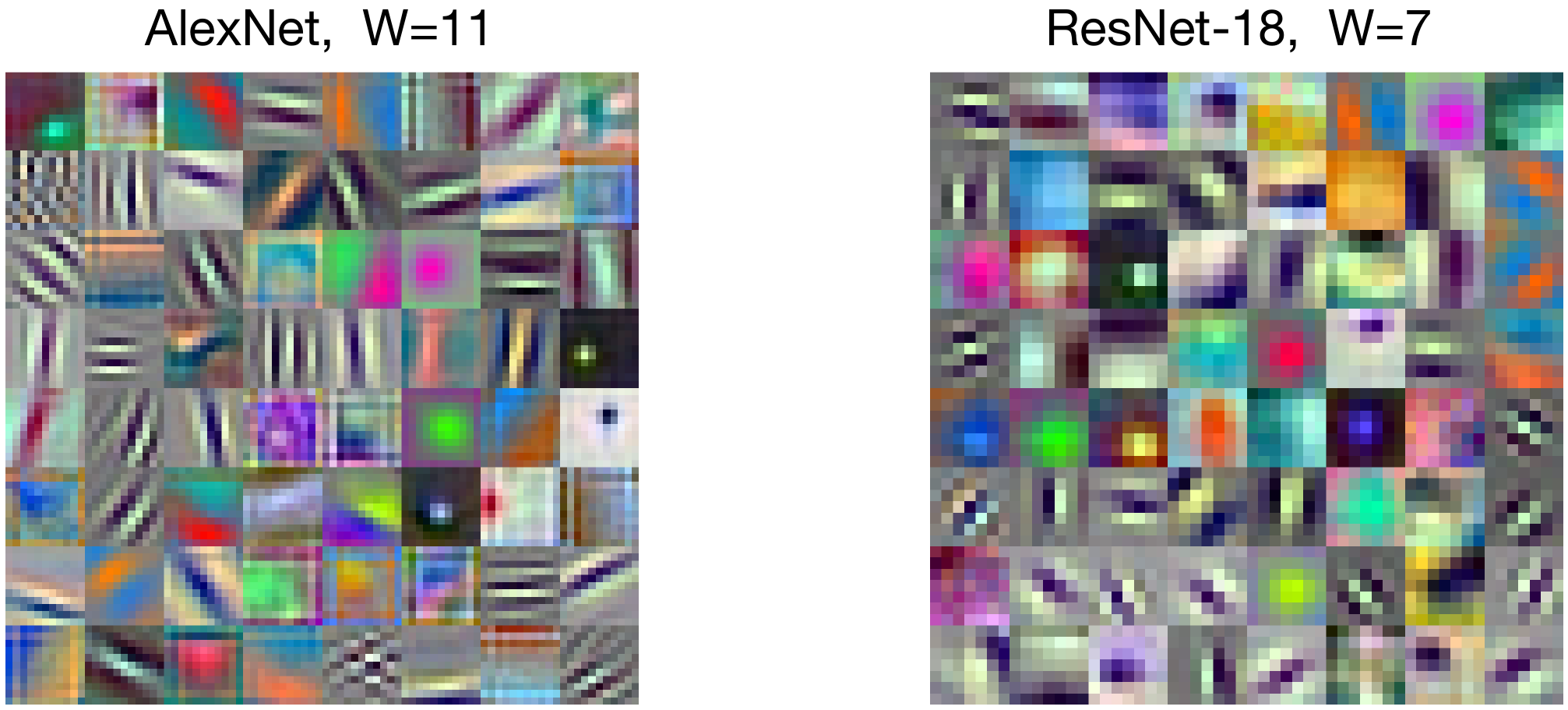}
\end{center}
\caption{\footnotesize{Convolutional filters from the first layer of two networks trained with backpropagation on ImageNet. AlexNet with $W=11$, and ResNet-18 with $W=7$. The same color code as in Fig. \ref{bio receptive fields}, \ref{ImageNet_rec_fields_fig} is used.  }}\label{Standard_Nets_fig}
\vspace{-0.2cm}
\end{figure} 
For completeness, in Fig. \ref{Standard_Nets_fig} we show filters learned by two standard networks used in computer vision: AlexNet \cite{AlexNet} and ResNet-18 \cite{ResNet}. These networks are much deeper than ours and were trained on the full resolution ImageNet dataset. These two aspects make it difficult to compare their filters with the ones learned by our algorithm.  However, they appear to look very different from the filters shown in Fig. \ref{bio receptive fields}, \ref{ImageNet_rec_fields_fig}.  

\section{Patch Normalization, Retinal Adaptation, and Shadows}
Natural scenes can have a several thousand fold range of light intensity variation \cite{Dunn}. At the same time, an 8-bit  digital camera has only 256 possible intensity values.  In order to cope with this huge variation of light intensities two separate systems exist in biological vision: a global control based on changing the size of the pupil, and a local adaptation on the retina. The latter, being the dominant one, enables the retina to have a good signal to noise ratio in both sun and shadow regions of a visual scene \cite{Dunn, Heeger}. 
\begin{figure}[h]
\begin{center}
\includegraphics[width = 0.85\linewidth]{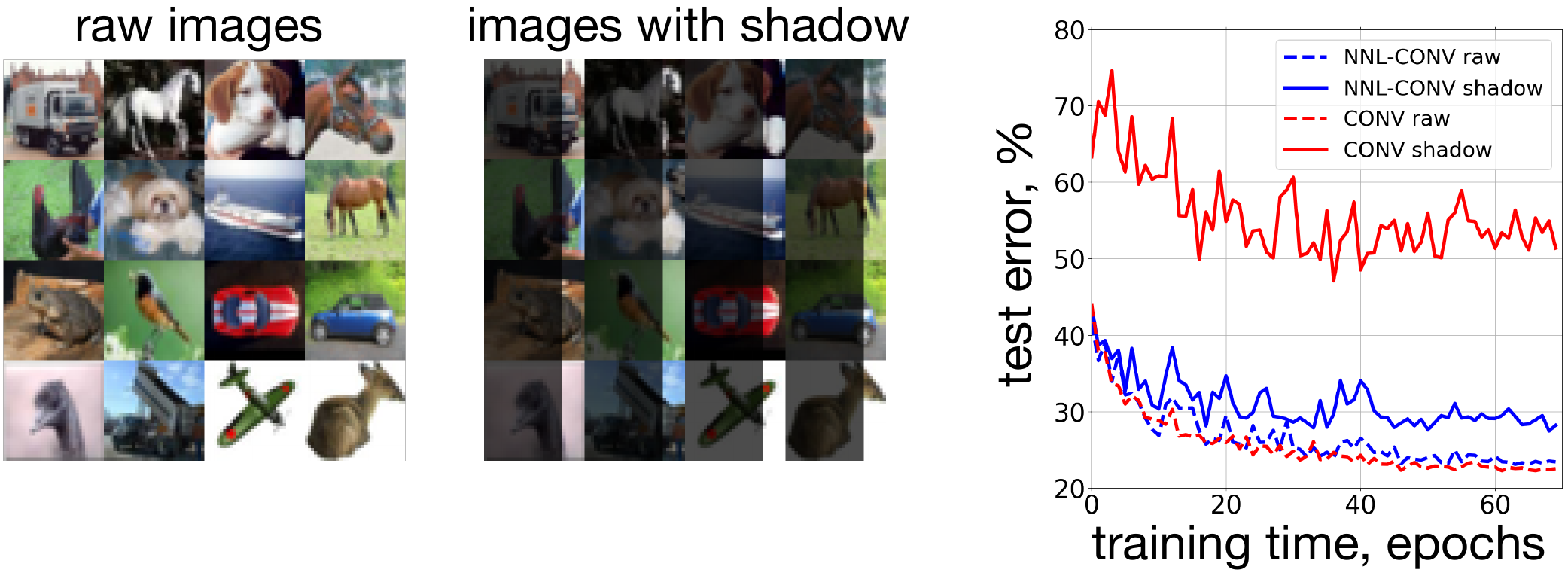}
\end{center}
\caption{\footnotesize{(Left) Randomly selected images from CIFAR-10 dataset. (Middle) Same images, first $25$ columns of each image are multiplied by an arbitrary parameter $0.3$ pixelwise. This imitates a shadow. (Right) Errors on the test set for the model with local patch normalization (blue) and a standard convolutional network (red). Raw images - dashed lines, shadowed images - solid lines. }}\label{shadows_fig}
\end{figure} 
Patch normalization, which is essential for a good performance of our algorithm, can be thought of as a mathematical formalization of the local circuitry on the retina that is responsible for this adaptation. Although we do not have a dataset with images of real scenes in various lighting conditions, it can be reasonably emulated  by multiplying images from CIFAR-10 dataset pixelwise by a function $I(x,y)$, which changes between zero and one. Patch normalization discards the overall normalization constant in every patch, which is strongly dependent on the light conditions, and focuses chiefly on the shape of an object that a given unit sees.

Examples of images constructed this way are shown in Fig. \ref{shadows_fig}, where $\approx 80\%$ of each image was covered by a shadow having $I(x,y)=0.3$. Human can see and correctly classify these ``shadowed'' images. Two networks: standard CONV net trained end-to-end, and NNL-CONV net trained as described in section \ref{section CIFAR} were trained on raw CIFAR-10 images, but tested on the shadowed images. Both networks had exactly the same architecture, shown in Fig.  \ref{networks errors} (middle). The results are shown in Fig. \ref{shadows_fig} (right).  While the errors on the raw CIFAR-10 images are approximately the same for the two networks ($\approx 23\%$), the error on the shadowed images of the NNL-CONV net is much lower ($\approx 28\%$) compared to the error of the standard CONV net ( more than $\approx 50\%$).  This illustrates that patch normalization can be a useful tool for dealing with images having large differences in light intensity (for example coming from shadows), without having images with these kinds of shadows in the training set.

A variety of intensity normalization schemes were used in the era of feature engineered systems, see for example \cite{Love, Collins}. Some of them, for example normalized cross correlation of \cite{Collins} or SIFT normalization \cite{Love}, have some similarities and differences with the proposed normalization of NNL-CONV neurons. The biggest difference between our approach and those feature engineered systems is that the filters used in our networks are learned and not hand-crafted.

\section{Transfer Learning}\label{transfer learning section}
The filters learned by our algorithm are independent of the task specified by the top layer classifier. This makes them natural candidates for the transfer learning. In order to test this idea we used the weights trained on ImageNet $32 \times 32$ as NNL-CONV filters in the architecture shown in Fig. \ref{networks errors} (middle) and retrained the top layer on the CIFAR-10 images. This gave an error $22.19\pm0.07\%$ (mean$\pm$ std over 5 training runs). This is more than $1\%$ lower than the error of the same network trained on CIFAR-10 images, which makes sense since ImageNet $32 \times 32$ has many more images than CIFAR-10 dataset.   Following the same procedure with filters obtain through end-to-end backpropagation training we obtained $22.32\pm0.09\%$. Thus, the locally trained filters perform as well as the standard ones, despite being learned by the local and unsupervised algorithm.

Transfer in the opposite direction - weights trained on CIFAR-10 used in the architecture shown in Fig. \ref{network_ImageNet} with top layer retrained on ImageNet $32 \times 32$ resulted in the top-1 error $85.38\%$, top-5 error $71.75\%$. This needs to be compared with top-1 error $84.13\%$, and top-5 error $70.00\%$ when trained using local learning directly on ImageNet $32 \times 32$. Again, having more images helps in reducing the error, but the difference is not very big.

\section{Computational Aspects}
The concept of local learning seems to be a powerful idea from the algorithmic perspective. Unlike end-to-end training, which requires keeping in memory the weights of the entire neural network together with activations of all neurons in all layers, in local training it is sufficient to keep in compute device memory (for example a GPU) only one layer of weights, a minibatch of data, and the activations of neurons only in that one layer.  Thus, local learning is appealing for the use on accelerators with a low memory capacity ($<16$GB). Additionally, since the weights are general, i.e. - derived directly from the data without the information about the task - it is possible to save and reuse them in different neural network architectures, without the need to retrain. This makes it possible to easily experiment with modular architectures (composed of blocks of different kinds of neurons), like the one shown in the middle panel of Fig. \ref{networks errors} and Fig. \ref{network_ImageNet}, without the need to recompute those weights multiple times.

The main limitation of the open source NumPy implementation \cite{bio_learning_github} is that it is sequential and not GPU accelerated. Thus it is extremely slow, which makes it unpractical for working with image datasets. At the same time, existing frameworks for deep learning are designed for end-to-end training. While it is certainly possible to implement the algorithm (\ref{learning rule}) using PyTorch or TensorFlow, such  implementations would not be performance optimal. Thus, we invested time in building a fast C++ library for CUDA that takes most advantage of the concept of local learning and optimizes the~performance.

Our fully parallel algorithm requires efficient (fast and programmable) hardware optimized mostly for i) matrix-matrix multiplications; ii) vector operations; iii) high-bandwidth connectivity between the processor and different types of memory; iv) sparse algebra to apply the effect of the activation function $g(\cdot)$. Parallel implementation of the activation function also requires the use of atomic operations. While the memory footprint of the input data we work with is in the range of 10GB to 400GB, the working data set (minibatch of data + model's  weights) requires between 30MB and 3GB of storage, which fits well into the V-100 GPU memory. We also use the Unified Virtual Address Space and the Address Translation Service \cite{P9_NPU} to access input data that can be spread over the GPU’s High Bandwidth Memory, the system (CPU) memory, the SSD device, and even the file system. The 50-75 GB/s [unidirectional] NVLink between each GPU and CPU makes it possible to do a fast data transfer from the CPU memory to the GPU when the input data does not fit to the GPU memory.

This results in a typical training time of filters on the ImageNet $32\times 32$ dataset (2,562,334 images) about 20-25 minutes, which is significantly faster than what would be possible with end-to-end training on a single GPU (typical training time for the architectures considered in this paper is of the order of 3-10 hours).

\section{Discussion and Conclusions}
This work is a proof of concept that local bottom-up unsupervised training is capable of learning useful and task-independent representations of images in networks with local connectivity. Similarly to \cite{Krotov_Hopfield_2019}, we focus on networks with one trainable hidden layer and show that their accuracies on standard image recognition tasks are close to the accuracies of networks of similar capacity trained end-to-end. The appealing aspects of the proposed algorithm are that it is very fast, conceptually simple (the weight update depends only on the activities of neurons connected by that weight, and not on the activities of all the neurons in the network), and leads to smooth weights with a well pronounced segregation of color and geometry features. We believe that this work is a first step toward algorithms that can train multiple layers of representations. Even if it turns out that this algorithm is only useful in the first layer, it can still be combined with backpropagation training in higher layers and take advantage of the smoothness of the first layer's weights. This might enhance interpretability of the networks or robustness against adversarial attacks, questions that require a comprehensive investigation.  

Convolutional neurons with patch normalization and steep activation functions, proposed in this paper, are essential for the good performance of our networks.  It is worth emphasizing that if the filters learned by our algorithm were simply substituted as weights in the conventional convolutional network, instead of the NNL-CONV network, the accuracy of the classification would be very poor. 

There exists a large family of unsupervised feature learning methods, such as learning with surrogate classes \cite{surrogate classes} , adversarial feature learning \cite{GAN, GAN_K_means}, etc. Local Hebbian learning and unsupervised learning with GANs are not mutually exclusive, but rather are complimentary. We hope that this work is a step toward merging these ideas and designing new powerful unsupervised learning algorithms.

\section*{Appendix. Technical Details of Experiments discussed in this paper.} 
We have done an extensive set of experiments varying various parameters of the local training algorithm. We have experimented with the following parameters of the convolutional blocks: size of the hidden layer $100 \leq K \leq 2000$, convolutional window $2 \leq W \leq 18$, strides $1 \leq ST \leq 4$, strength of the anti-Hebbian learning $0 \leq \Delta  \leq 0.3$, etc. Additionally, we have experimented with the hyperparameters of the full architecture: pooling size $ 1 \leq W_p \leq 18$, power $1 \leq n \leq 100$ (this parameter was varied with increment $10$), pooling strides $1 \leq ST_p \leq 4$, size of the minibatch, learning rate annealing schedule.

All the parameters were determined on the validation set, as discussed in the main text for both the networks trained locally and the networks trained end-to-end.

For the experiments reported in Fig. \ref{networks errors} (left network),  the optimal hyperparameters were: $m=2$, $\Delta = 0.2$, $K=400$, $W=4$, $n=40$, $W_p=11$, convolutional stride $ST=1$, pooling stride $ST_p=2$, minibatch size for local training was $1000$ patches, minibatch size for the top layer backpropagation training was $300$ images.  The convolutional filters were trained for $500$ epochs with learning rate linearly decreasing from $\varepsilon_0=1 \cdot 10^{-4}$ to zero. The top layer was trained for $70$ epochs with the following schedule of the learning rate decrease: 
\begin{equation}
\varepsilon= \left\{ 
\begin{array}{cl}1\cdot 10^{-4}, & \text{epoch}\leq 15\\ 
8 \cdot 10^{-5}, & 15<\text{epoch}\leq 30 \\
5 \cdot 10^{-5}, & 30<\text{epoch}\leq 45 \\
2 \cdot 10^{-5}, & 45<\text{epoch}\leq 60 \\
1 \cdot 10^{-5}, & 60<\text{epoch}\leq 70
\end{array}\right.\label{learning rate annealing CIFAR}
\end{equation}

The backpropagation counterpart of this network was chosen to have (almost) the same capacity as the NNL-CONV network. Standard CONV filters have biases as parameters of the convolutional filters, while in NNL-CONV nets these biases are set to zero. Thus, strictly speaking the CONV net always has a little bit more parameters than the corresponding (with the same number of channels $K$ and the same set of windows $W$) NNL-CONV net, but this difference is equal to the number of convolutional filters, and thus is much smaller than the total number of parameters of the network. We ignore this small difference and assume that the two networks have the same capacity.

The end-to-end counterpart of the NNL-CONV network shown in Fig. \ref{networks errors} (left) had the following hyperparameters: $K=400$, $W=4$, $ST=1$, $ST_p=2$, minibatch size was $300$ images, and the activation function was a ReLU. The network was trained for $70$ epochs using the learning rate decrease (\ref{learning rate annealing CIFAR}). In order to make the comparison of the two networks simpler we used the same value of $W_p=11$ for this network, instead of optimizing it on the validation set. We have also checked that additionally optimizing over the parameter $W_p$ on the validation set does not significantly change the accuracy on the test set. The network trained end-to-end finds solutions of approximately the same accuracy for a broad range of $W_p$ around the optimum.

For the NNL-CONV net shown in Fig. \ref{networks errors} (middle) the remaining hyperparameters (not shown in the figure) were the following. The sequence of $\Delta$ for the five blocks was: $\Delta=[0.1,\ 0.1,\ 0.2,\ 0.15,\ 0.2$. For all blocks $m=2$, $ST=1$, $ST_p=2$, minibatch size for the local training was 1000 patches. The learning rate linearly decreased from $\varepsilon_0=1 \cdot 10^{-4}$ to zero during $500$ epochs. The minibatch size for the top layer training was $300$ and the learning rate annealing followed the schedule (\ref{learning rate annealing CIFAR}).

For the network trained end-to-end, minibatch size was 300 images, the set of pooling windows was: $W_p=[14,11,11,7,11]$ (not optimized on the validation set). We checked that this additional optimization would not change the results of Fig. \ref{networks errors}.  Activation functions were ReLU.  The learning rate annealing followed (\ref{learning rate annealing CIFAR}).

For the experiments reported in Fig. \ref{network_ImageNet} the remaining hyperparameters (not shown in the figure) were the following. The sequence of $\Delta$ for the four blocks were: $\Delta=[0.1,\ 0.2,\ 0.2,\ 0.2]$. For all blocks $m=2$, $ST=1$, $ST_p=2$, minibatch size for the local training was 10000 patches. The learning rate linearly decreased from $\varepsilon_0=1 \cdot 10^{-4}$ to zero during $50$ epochs. The minibatch size for the top layer training was $200$ and the learning rate annealing followed the schedule
\begin{equation}
\varepsilon= \left\{ 
\begin{array}{cl}1\cdot 10^{-4}, & \text{epoch}\leq 15\\ 
8 \cdot 10^{-5}, & 15<\text{epoch}\leq 25 \\
5 \cdot 10^{-5}, & 25<\text{epoch}\leq 35 \\
2 \cdot 10^{-5}, & 35<\text{epoch}\leq 45 \\
1 \cdot 10^{-5}, & 45<\text{epoch}\leq 48
\end{array}\right.\label{learning rate annealing ImageNet}
\end{equation}
for $48$ epochs.

For the network trained end-to-end the training was done for $5$ epochs with the learning rate $1\cdot 10^{-4}$ with minibatch of size 200 images. The same set of pooling windows as specified in Fig. \ref{network_ImageNet} was used. Additional optimization over these pooling windows does not result in improved accuracy.

For the experiments in reported in Fig. \ref{shadows_fig} the same settings were used as discussed above for the network shown in Fig. \ref{networks errors} (middle) and its end-to-end trained counterpart. The only difference is that in addition to testing the accuracy on raw images, the accuracy was also tested on shadowed images.

For the experiments reported in section \ref{transfer learning section} the setting corresponding to the setting for the target (to which the transfer is made) network was used. The set of $\Delta$ for the weights trained on ImageNet $32\times 32$ (and transferred to CIFAR-10) was $\Delta=[0,\ 0.1,\ 0.2,\ 0.2,\ 0.2]$ for the five blocks. Parameter $m$ was set to $m=2$ for all the five blocks, minibatch size for local training was $10000$ patches.  The convolutional filters were trained for $50$ epochs with learning rate linearly decreasing from $\varepsilon_0=1 \cdot 10^{-4}$ to zero. 
The set of $\Delta$ for the weights trained on CIFAR-10 (and transferred to ImageNet $32\times 32$) was $\Delta=[0.1, \ 0.2,\ 0.15,\ 0.2]$ for the four blocks. Parameter $m$ was set to $m=2$ for all the four blocks, minibatch size for local training was $1000$ patches.  The convolutional filters were trained for $500$ epochs with learning rate linearly decreasing from $\varepsilon_0=1 \cdot 10^{-4}$ to zero.

No dropout, noise injection, data augmentation or data preprocessing of any kind were used in this paper. We leave the investigation of the influence of these methods on the accuracy of our algorithm for a separate study. 

\section*{Acknowledgements} We thank Quanfu Fan and Hilde Kuehne for useful discussions.

\end{document}